\documentclass[runningheads]{llncs}

\usepackage{eccv}

\usepackage{eccvabbrv}

\usepackage{graphicx}
\usepackage{booktabs}
\usepackage{multirow}
\usepackage{pifont}

\newcommand{\cmark}{\ding{51}}
\newcommand{\xmark}{\ding{55}}

\usepackage[accsupp]{axessibility}  

\usepackage{hyperref}

\usepackage{orcidlink}
\usepackage[table]{xcolor}

\definecolor{cbblue}{RGB}{0,114,178}
\definecolor{cborange}{RGB}{230,159,0}
\definecolor{cbgreen}{RGB}{0,158,115}
\definecolor{cbpurple}{RGB}{204,121,167}

\usepackage[x11names]{xcolor}
\usepackage{tikz}
\usepackage{tabularx}
\usepackage{array}

\newcommand{\cbox}[1]{\textcolor{#1}{\rule{0.5em}{0.5em}}}
\usepackage[ruled,vlined,linesnumbered]{algorithm2e}
\usepackage{multicol}
\usepackage{bm}

\begin{document}

\title{Can You Hear, Localize, and Segment Continually? An Exemplar-Free Continual Learning Benchmark for Audio-Visual Segmentation}


\author{Siddeshwar Raghavan\inst{1}\orcidlink{0000-0002-3079-2585} \and
Gautham Vinod\inst{1}\orcidlink{0000-0002-2488-2541} \and
Bruce Coburn\inst{1}\orcidlink{0009-0004-5284-6718} \and
Fengqing Zhu\inst{1}\orcidlink{0000-0002-3863-3220}
}

\authorrunning{S Raghavan et al.}

\institute{Purdue University, West Lafayette, IN 47906, USA \\
\email{\{raghav12, gvinod, coburn6, zhu0\}@purdue.edu}}

\maketitle

\begin{abstract}
Audio-Visual Segmentation (AVS) aims to produce pixel-level masks of sound producing objects in videos, by jointly learning from audio and visual signals. However, real-world environments are inherently dynamic, causing audio and visual distributions to evolve over time, which challenge existing AVS systems that assume static training settings. To address this gap, we introduce the first exemplar-free continual learning benchmark for Audio-Visual Segmentation, comprising four learning protocols across single-source and multi-source AVS datasets. We further propose a strong baseline, ATLAS, which uses audio-guided pre-fusion conditioning to modulate visual feature channels via projected audio context before cross-modal attention. Finally, we mitigate catastrophic forgetting by introducing Low-Rank Anchoring (LRA), which stabilizes adapted weights based on loss sensitivity. Extensive experiments demonstrate competitive performance across diverse continual scenarios, establishing a foundation for lifelong audio-visual perception. Code is available at${}^{*}$\footnote{Paper under review} - \hyperlink{https://gitlab.com/viper-purdue/atlas}{https://gitlab.com/viper-purdue/atlas} 
  
\end{abstract}

\section{Introduction}

\begin{figure}[ht!]
    \centering
    \includegraphics[width=1\linewidth]{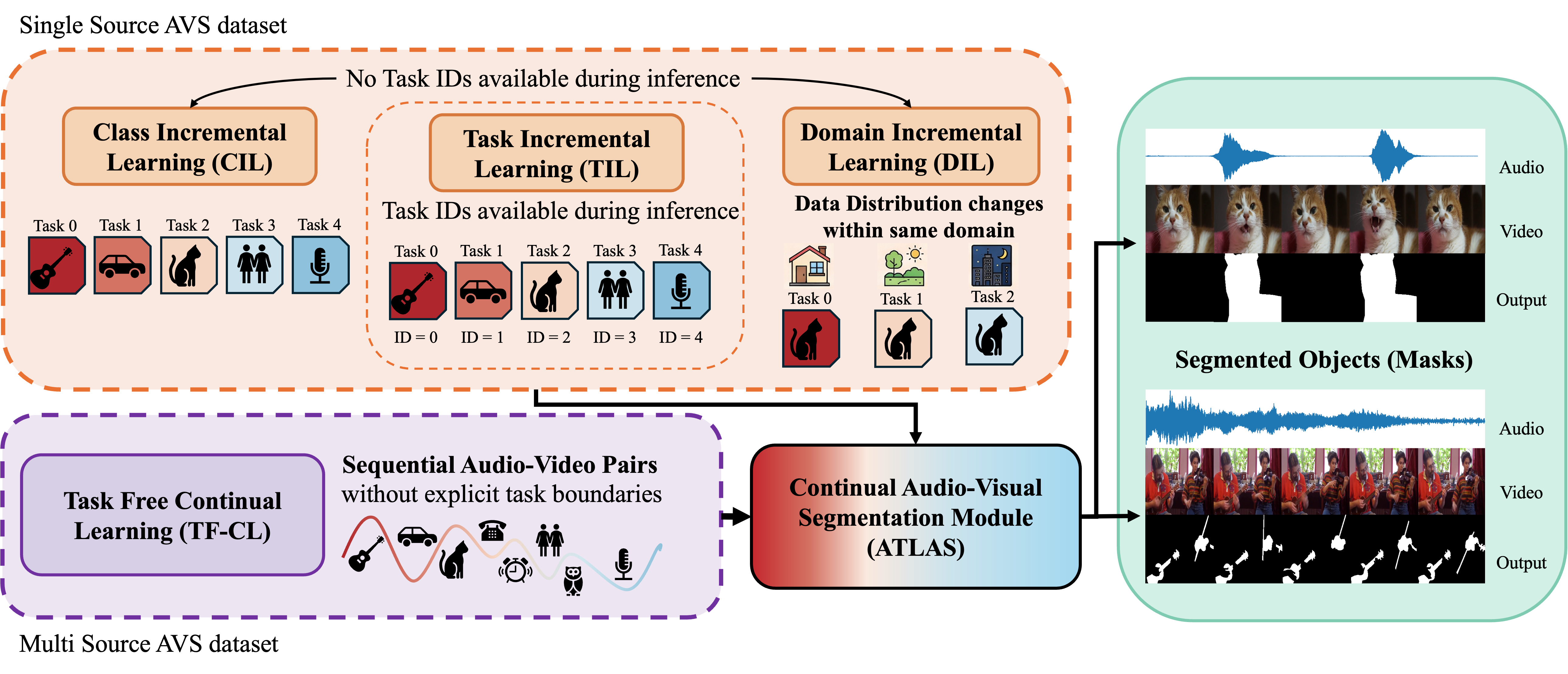}
    \caption{ An overview of Exemplar Free Continual Learning Benchmark for Audio-Visual Segmentation}
    \label{fig:cl-avs-intro}
\end{figure}

Humans can naturally recognize and localize objects by jointly interpreting visual cues and the sounds they produce through everyday experience~\cite{Leavitt2011-hg, Newell2023, Mihalik6600}. This capability is fundamental to how we perceive, understand, and interact with complex environments. Audio-Visual Segmentation (AVS) ~\cite{zhou2022avs, zhou2024avss} targets this challenge by producing a pixel level map of the objects that produce sounds in the video frames. However, humans continuously adapt to changing surroundings, acquiring new visual and auditory patterns while retaining previously learned knowledge. Hearing a new instrument for the first time, for instance, does not cause listeners to forget the sound of a dog's bark. Replicating this lifelong learning ability in machines remains highly challenging due to the intricate correspondence and interdependence between audio and visual modalities, the scarcity of precise localization supervision, and the persistent issue of catastrophic forgetting~\cite{cf_1,cf_2,cf_3}. 

Since its introduction, AVS has attracted considerable attention, with methods exploring audio-visual correspondence through transformer-based fusion~\cite{zhou2024avss}, bidirectional cross-modal attention~\cite{AVS_birdir}, and temporal audio-visual pixel interaction strategies~\cite{zhou2022avs}. Despite these advances, a critical direction remains under-explored, as real-world audio-visual perception is inherently \emph{continual}. In practice, a deployed model encounters new sound categories over time, ranging from musical instruments and animals to vehicles and human speech. The model must learn them without retraining from scratch or revisiting prior data. Exemplar Free Continual Learning (EFCL) provides a principled framework for this setting, enabling models to acquire new knowledge sequentially while mitigating catastrophic forgetting~\cite{cf_1,cf_2,cf_3} without storing any data. The emergence of Pre-Trained Models (PTMs) has further accelerated progress in this direction, as EFCL methods built on strong pre-trained representations~\cite{L2P, ranpac, fecam, DGR, raghavan2025pandapatchdistributionaware} have demonstrated substantially better real-world applicability compared to their trained-from-scratch frameworks. While Continual Learning (CL) was originally studied in the context of image classification~\cite{LWF, aljundi2018memoryawaresynapseslearning, synaptic_CL, EWC,sraghavan2024delta,Raghavan_2024_WACV}, the field has since expanded considerably, including continual semantic segmentation~\cite{CL_semanticseg}, multi-modal continual learning~\cite{yu2024recentadvancesmultimodalcontinual}, and continual reinforcement learning~\cite{pan2025surveycontinualreinforcementlearning}, reflecting a broader shift toward more realistic and open-ended learning scenarios.

However, directly applying these methodologies to the EFCL setting or re-purposing existing methods to the EFCL setting is not trivial. AVS requires the model to simultaneously maintain cross-modal alignment between audio and visual streams, preserve fine-grained spatial segmentation boundaries, and retain previously learned sound producing object associations, all without revisiting or storing past data. The multimodal nature of AVS amplifies forgetting. Degradation in either modality or in their cross-modal alignment can cause failure, even if each modality retains useful information. Beyond forgetting, continual AVS faces several additional challenges. First, the introduction of new sounding categories causes \emph{cross-modal interference}, where the feature spaces of both the audio and visual encoders must shift to accommodate new patterns, risking misalignment between modalities that were previously well-correlated. Second, the \emph{evaluation complexity} increases because continual AVS requires dense pixel-level metrics that evaluate segmentation quality and audio-visual alignment while accounting for continual learning across previously encountered tasks.

To bridge this gap, we make three contributions. \textbf{First}, we introduce the exemplar-free continual learning benchmark for Audio-Visual Segmentation (CL-AVS), illustrated in Figure~\ref{fig:cl-avs-intro}. The benchmark evaluates four continual learning protocols across two AVS datasets: Task-Incremental (TIL), Class-Incremental (CIL), and Domain-Incremental Learning (DIL) on the Single Source AVS (SS-AVS) dataset~\cite{zhou2022avs}, and Task-Free Continual Learning~\cite{task_free_cl} on the Multi-Source AVS (MS-AVS) dataset~\cite{zhou2022avs}. \textbf{Second}, we propose \emph{ATLAS}, an exemplar-free baseline built on LoRA~\cite{lora} adapters for parameter-efficient continual learning. It introduces an audio guided pre-fusion conditioning prior to audio–visual fusion and a Low-Rank Anchoring (LRA) mechanism to mitigate parameter drift across tasks. \textbf{Third}, we conduct extensive experiments across diverse continual learning algorithms and AVS frameworks, with detailed analysis highlighting the challenges of continual AVS and the effectiveness of our baseline

\section{Related Works}
\label{sec:relwork}
 
\subsection{Audio-Visual Segmentation}
Audio-Visual Segmentation (AVS) aims to localize sound producing objects in video frames through pixel-level masks by jointly modeling audio and visual signals. The task was introduced in AVSBench~\cite{zhou2022avs}, which established a benchmark for binary audio-visual segmentation, and later extended to semantic AVS with category-aware predictions~\cite{zhou2024avss}. Early approaches typically adopt encoder–decoder architectures that fuse visual representations with audio embeddings using attention or correlation mechanisms. While effective in controlled environments, these models often struggle in realistic scenes containing multiple sound sources, background noise, and visually salient but silent objects~\cite{chen2023closer_avs}.

Recent AVS research broadly follows two design paradigms. \textbf{Vision-anchored} approaches leverage strong visual priors and treat audio as a conditioning signal, commonly using transformer-based fusion or vision foundation models to improve spatial precision and boundary quality~\cite{huang2025_visual_centric_avs, mo2023_av_sam, li2025_robust_avs}. Although effective, heavy reliance on visual cues may lead to incorrect segmentation when audio–visual correspondence weakens. In contrast, \textbf{audio-anchored} methods emphasize sound-driven localization through bidirectional interaction~\cite{AVS_birdir} or audio-conditioned attention mechanisms~\cite{chen2023_bootstrap_avs}, improving robustness when visual motion is ambiguous but introducing sensitivity to noisy audio. To reduce annotation cost, several works further explore weakly supervised and self-supervised learning by exploiting natural audio–visual correspondence or pseudo-label generation~\cite{mo2023ws_avs, liu2024annotation_free_avs}.

Despite rapid progress, existing AVS methods assume static training distributions where all categories are jointly available. In real-world deployment, however, audio-visual systems encounter continuously evolving environments with new sounding categories appearing over time. To bridge this gap and set up the foundation we propose an exemplar free continual learning benchmark for AVS.

\subsection{Continual Learning}
Continual Learning (CL) studies learning from sequentially arriving data while mitigating catastrophic forgetting~\cite{cf_1, cf_2, cf_3}. Early work largely focused on image classification, exploring replay-based, regularization-based, and parameter-isolation strategies~\cite{LWF, EWC, synaptic_CL, aljundi2018memoryawaresynapseslearning}. With the emergence of large pretrained models and increasing privacy constraints, recent research has shifted toward replay-free adaptation built on pretrained representations~\cite{ranpac, L2P, DGR, fecam}, enabling scalable continual learning without storing past data.

Beyond classification, CL has expanded to dense prediction problems such as continual semantic segmentation~\cite{chen2024strike, EIR_2025, chen2025csta, Dong2025HVPL, CL_semanticseg} and to multimodal learning settings~\cite{yu2024recentadvancesmultimodalcontinual, raghavan2025pandapatchdistributionaware}. Extending continual learning to audio-visual segmentation introduces additional challenges, as models must maintain class-level knowledge and cross-modal alignment while making fine-grained spatial predictions.

\section{Methodology}
In this section, we present the datasets, problem formulation for Continual Audio-Visual Segmentation, introduce the benchmark protocols defining CL-AVS, and finally describe our proposed methodology ATLAS for addressing this challenge.

\subsection{Datasets}
To provide the necessary context for our formal problem statement, we first describe the datasets used in this work. We utilize the datasets proposed in AVSBench~\cite{zhou2022avs}, specifically the Single-Source AVS (SS-AVS) and Multi-Source AVS (MS-AVS) datasets. The SS-AVS consists of 4,932 videos across train, test, and validation splits with 23 categories. It is defined in a semi-supervised setting, where only the first frame of each video has ground-truth pixel level segmentation. In contrast, the MS-AVS dataset contains 424 videos covering the same 23 categories. In this setting, the videos are indexed by video ID rather than class labels, since each video clip may include multiple sound-producing objects. MS-AVS is fully supervised, with pixel-level annotations available for every frame, but class labels are unavailable. This fundamental difference between the two datasets directly motivates the design of our continual audio-visual segmentation (CL-AVS) benchmark protocols.

\subsection{Problem Formulation}
The goal of audio-visual segmentation (AVS) is to learn a model capable of partitioning video frames into a set of regions that localize sound-producing objects given the corresponding audio signal. These tasks are distinguished by the details of the predicted masks. Binary AVS produces a set of foreground masks separating sounding objects from the background, while semantic AVS additionally assigns each sounding region a category label over a predefined set of object classes. Now we provide a general formulation for continual learning in AVS (CL-AVS) that unifies both settings, extending continual segmentation frameworks~\cite{cermelli2023comformer, plop} to the audio-visual domain.

CL-AVS trains the model over a sequence of $N$ tasks ($T_1, \dots, T_N$), each introducing a new set of sound producing object categories. At each learning step $k$, a uniform number of $M$ audio-video pairs are sampled from the dataset $\mathcal{D}$ to construct the task $T_k = \{(V_i^{(k)}, A_i^{(k)}, Y_i^{(k)})\}_{i=1}^{M}$. Here, each video is represented as a sequence of frames $V_i^{(k)} = \{v_{i,t}^{(k)}\}_{t=1}^{C_i}$ with temporally aligned audio segments $A_i^{(k)} = \{a_{i,t}^{(k)}\}_{t=1}^{C_i}$, where $C_i$ denotes the number of frames for the $i$-th video.
The ground-truth annotation $Y_i^{(k)} = \{y_{i,t}^{(k)}\}_{t=1}^{C_i}$ varies according to the task formulation. Let $\mathcal{K}^k$ denote the set of sounding object classes introduced at step $k$, and $\mathcal{K}^{1:k} = \bigcup_{j=1}^{k} \mathcal{K}^j$ the cumulative set of all classes seen up to step $k$, then,

$$y_{i,t}^{(k)} = 
\begin{cases}
m_{i,t}^{(k)} \in \{0,1\}^{H \times W}  \quad \quad & \text{\textbf{binary setting}}\\
s_{i,t}^{(k)} \in {\mathcal{K}^{1:k} \cup \{0\}}^{H \times W} \quad \quad & \text{\textbf{semantic setting}}
\end{cases}
$$

\subsection{Continual Learning Settings}
There are three major settings in Continual Learning, namely Task Incremental, Class Incremental, and Domain Incremental~\cite{lamda_pilot}. We adapt these settings to our benchmark CL-AVS on Single Source AVS (SS-AVS) dataset~\cite{zhou2022avs} and additionally extend a Task Free~\cite{blurryCL} protocol suited to Multi-Source AVS (MS-AVS) dataset~\cite{zhou2022avs}. 

\paragraph{\textbf{Task Incremental AVS:}}
In this setting, a new set of sound producing object classes $\mathcal{K}^k$ are introduced at each learning step $k$. A task identifier (ID) is provided at both training and test time, indicating which step a given sample belongs to. At test time, the model is evaluated on all seen classes $\mathcal{K}^{1:k}$ using the corresponding task identifiers.

\paragraph{\textbf{Class Incremental AVS:}}
Similar to the task incremental setting, new sound producing object classes $\mathcal{K}^k$ are introduced at each step $k$. However, no task identifier is available at test time. The model must therefore differentiate all previously seen classes $\mathcal{K}^{1:k}$ without knowledge of which step a given sample originated from.

\paragraph{\textbf{Domain Incremental AVS:}}
In this setting, the model operates within a single sounding object category (e.g., \textit{barking dog}) and learns sequentially across videos of the same class but with varying data distributions, such as different visual appearances, scenes, or audio conditions. Formally, the class set remains fixed across all steps, i.e., $\mathcal{K}^1 = \mathcal{K}^2 = \dots = \mathcal{K}^N$, while the underlying data distribution of $\mathcal{D}_k$ shifts at each step. The model is evaluated on its ability to retain segmentation performance across all encountered distributions without forgetting earlier ones.

\paragraph{\textbf{Task-Free Continual AVS:}}
We extend the task-free CL paradigm~\cite{blurryCL} to the multi-source AVS dataset~\cite{zhou2022avs}, where explicit class labels are unavailable. The model observes a stream of videos across $N$ tasks with blurry boundaries. The objective is to perform binary segmentation (sounding vs.\ non-sounding) without semantic distinction, evaluated over many tasks (e.g., $N{=}50$) under the anytime inference principle~\cite{blurryCL}.

\subsection{ATLAS Framework}
\begin{figure}[ht!]
    \centering
    \includegraphics[width=1\linewidth]{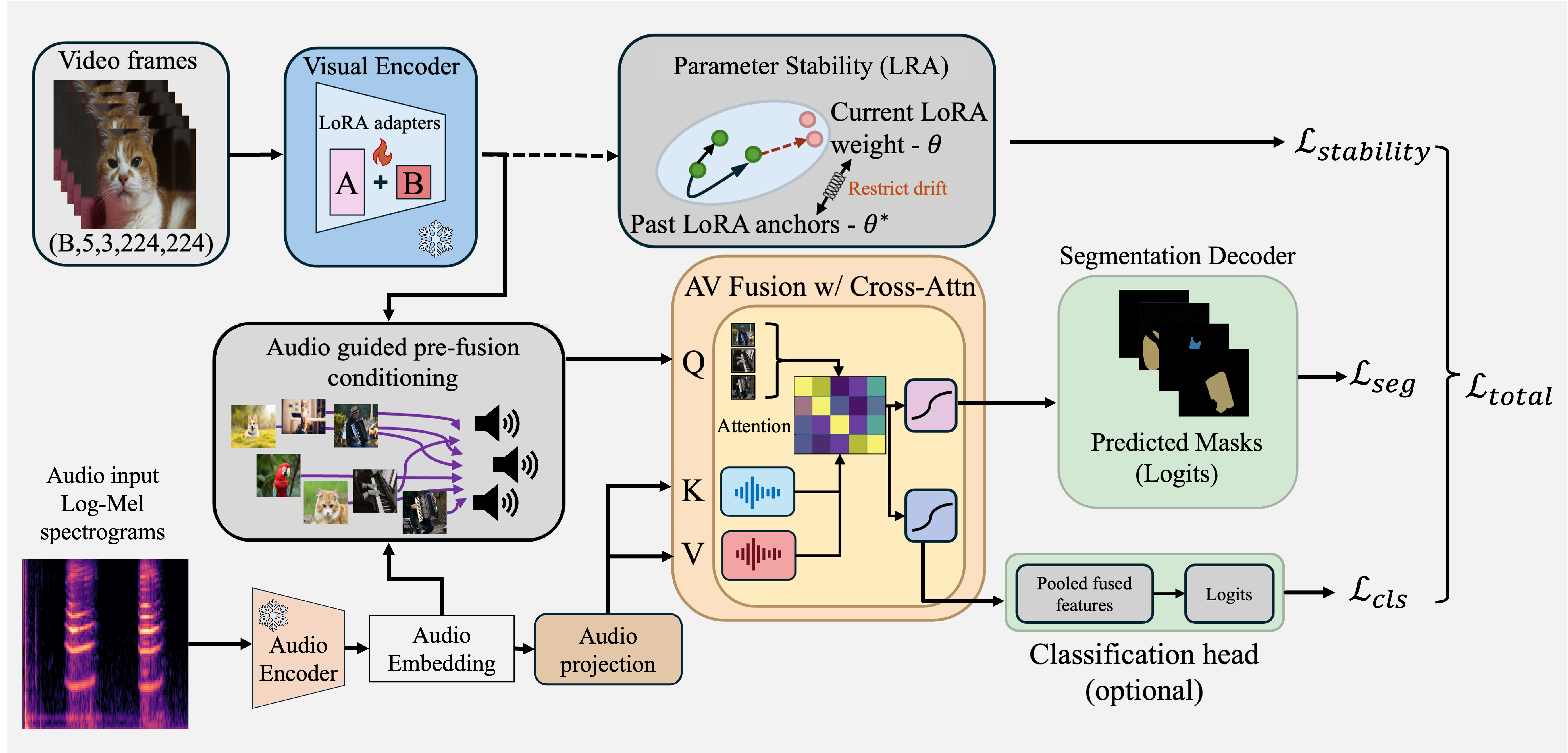}
    \caption{\textbf{Overview of ATLAS:} The framework performs exemplar-free continual audio-visual segmentation using frozen encoder backbones with parameter-efficient LoRA adapters in the visual pathway. Audio-guided pre-fusion conditioning is followed by cross-attention, where conditioned visual features act as queries and audio features serve as keys and values. To mitigate catastrophic forgetting across tasks, our Low-Rank Anchoring (LRA) module dynamically restricts LoRA parameter drift. The fused representation is decoded for segmentation with an optional classification branch. }
    \label{fig:atlas}
\end{figure}
We introduce Adaptive Task Learning with Anchored Stability (ATLAS), shown in Figure~\ref{fig:atlas}, an exemplar-free baseline for continual audio-visual segmentation. It segments sound-producing objects from video frames across sequential tasks without retraining or revisiting past data. Following the problem formulation, at a given task $T_k = \{ (V_i^{(k)},A_i^{(k)},Y_i^{(k)})\}_{i=1}^M$, ATLAS maps each pair $(V_i^{(k)},A_i^{(k)})$ to mask logits and class logits (optionally) while constraining parameter drift across tasks to minimize catastrophic forgetting during continual learning. Let the visual encoders and audio encoders be $\Phi(\cdot;\Theta_v)$ and $\Phi(\cdot;\Theta_a)$, respectively. To enable parameter efficient adaptation, we train selected linear maps in the visual encoder backbone with LoRA~\cite{lora} adapters. For each pretrained matrix $W_0$, the updated weight is defined as:
\begin{equation}
\label{eqn:lora}
    W = W_0 + \Delta W \textrm{  and  }\Delta W = \frac{\alpha}{r}BA
\end{equation}
where $A \in \mathbb{R}^{r \times d_{in}}$ and $B \in \mathbb{R}^{d_{out} \times r}$ with rank $r \ll min(d_{in}, d_{out})$. The hyperparameter $r$ acts as a constant scaling factor. The LoRA adapter effectively update the weights of the visual encoder to produce adapted visual features. For each sample $i$, the visual and audio encoded features are
\begin{equation}
    X_{v, i}^{(k)} = \Phi_v(V_i^{(k)}; W) \textrm{  and  } X_{a, i}^{(k)} = \Phi_a(A_i^{(k)}; \Theta_a)
\end{equation}
These visual encoded features then flow into the \textbf{Audio Guided Pre-Fusion Conditioning} module which injects global audio context into visual tokens via learned channel wise modulation $\bar{X}_{v,i}^{(k)} = \mathcal{C}(X_{v, i}^{(k)},X_{a, i}^{(k)})$.
The audio-guided conditioning operator is $\mathcal{C(\cdot)}$ (detailed steps provided in supplementary material). Intuitively, this learned channel-wise modulation acts as a feature-level gating mechanism. Here, the audio features are projected to the visual token space to produce modulation parameters and conditioned tokens $\bar{X}_{v,i}^{(k)}$. By projecting the audio context into scaling and shifting parameters, the network selectively amplifies the specific visual channels that correspond to the sound-producing object while suppressing irrelevant background noise. Before AV fusion, this preliminary step aligns the visual features with sound relevant regions. These pre-conditioned visual features in turn serve as the queries (Q) in the subsequent \textbf{AV Fusion with Cross-Attention} module. We compute the Q, K, and V from these pre-conditioned visual features and the audio features respectively:
\begin{equation}
Q_i^{(k)} = W_Q\bar{X}_{v,i}^{(k)},\quad
K_i^{(k)} = W_KX_{a,i}^{(k)},\quad
V_i^{(k)} = W_VX_{a,i}^{(k)}.
\end{equation}
To preserve the spatial integrity of the visual features during fusion, the fused features are obtained by taking softmax over the scaled dot product attention and applying a residual connection with Layer Normalization~\cite{attentionallyouneed}:
\begin{equation}
    F_i^{(k)} = \texttt{LayerNorm}\left(\bar{X}_{v,i}^{(k)} + \texttt{softmax}\left(\frac{Q_i^{(k)}K_i^{(k)^T}}{\sqrt{d}}\right)V_i^{(k)} \right)
\end{equation}
The decoder $\mathcal{D}$, which also contains the LoRA adapted weights to mitigate catastrophic forgetting, projects these fused multimodal features to the resolution of the original video frames to output the mask logits:
\begin{equation}
    \hat{M}_i^{(k)} = \mathcal{D}\left(F_i^{(k)} \right)
\end{equation}
Additionally, to support the prediction of class logits, a global pooling layer followed by a linear classification head $\mathcal{H}$ processes the combined representation to output the class logits
\begin{equation}
    \hat{z}_i^{(k)} = \mathcal{H}\left(F_i^{(k)} \right)
\end{equation}
For the binary setting in our CL-AVS benchmark, $\left(y_{i,t}^{(k)} = m_{i,t}^{(k)} \right)$, segmentation is optimized using a combination of Binary Cross Entropy and Dice~\cite{dice} losses to handle severe foreground to background imbalances, in a supervised manner:
\begin{equation}
\label{eqn:seg}
    \mathcal{L}_{seg}^{(k)} = \frac{1}{M}\sum_{i=1}^{M}\left[ \mathcal{L}_{BCE}\left( \hat{M}_{i}^{(k)}, m_{i}^{(k)}\right)+ \mathcal{L}_{Dice}\left( \hat{M}_{i}^{(k)}, m_{i}^{(k)}\right)\right ]
\end{equation}
And for the semantic setting where we predict the logits in a supervised way over the classes seen so far $Q^{1:k}$, the classification term is standard Cross Entropy (CE):
\begin{equation}
\label{eqn:cls}
    \mathcal{L}_{cls}^{(k)} = \frac{1}{M}\sum_{i=1}^{M}\texttt{CE}\left( \hat{z}_i^{k}, c_i^{(k)} \right)
\end{equation}
where $c_i^{(k)} \in Q^{1:k}$ is the sample level class label. Finally, we introduce the \textbf{Low-Rank Anchoring (LRA)} module to minimize drift between the past anchor weights and current weights to mitigate catastrophic forgetting. Let $\theta$ denote the current trainable parameters, $\theta^*$ the previous task's anchor weights, and $\Omega_i$ the parameter importance weight. Rather than computing static Fisher approximations, $\Omega_i$ is computed dynamically during training by accumulating the product of parameter gradients and their updates. This effectively tracks loss sensitivity along the optimization trajectory. The stability regularization is applied to LoRA matrices and decoder weights. The stability term is defined as:
\begin{equation}
\label{eqn:stab}
    \mathcal{L}_{stab} = \frac{c}{2}\sum_{i}^{}\Omega_i\left( \theta_i - \theta_i^* \right)^2
\end{equation}
where $c$ is a hyperparameter controlling the regularization strength. Our final objective at any task $k$ is the combination of losses defined in equations~\ref{eqn:seg},\ref{eqn:cls},\ref{eqn:stab}
\begin{equation}
    \mathcal{L}_{total}^{(k)} = \mathcal{L}_{seg}^{(k)} + \lambda_{cls}\mathcal{L}_{cls}^{(k)} + \mathcal{L}_{stab}^{(k)}
\label{eqn:tot_loss}
\end{equation}

\section{Experimentation Setup}

\subsection{Implementation Details}
To ensure rigorous evaluation, we select representative AVS and continual learning methods, prioritizing publicly available code and strong reported performance. For non-public implementations (marked with $*$), we reconstructed models using LLM-assisted development (Anthropic Claude Code~\cite{claude} and Google Gemini~\cite{gemini}).

Our experimental setup spans four closely related categories of methods adaptable to end-to-end CL-AVS training:
(1) (\cbox{cbblue}) Exemplar-free continual learning methods adapted to AVS by replacing (or adding) classification heads with segmentation heads along with audio encoder;
(2) (\cbox{cborange}) Static AVS models trained with continual data streams;
(3) (\cbox{cbpurple}) continual semantic segmentation frameworks extended with audio encoders and MSE-based cross-modal alignment; and
(4) (\cbox{Bisque2}) the replay-based CL-AVS baseline~\cite{CMR}.

Across our experiments we use use PANN~\cite{pann} (trained on AudioSet~\cite{audioset}) as the audio encoder. Visual backbones follow their original designs: most methods, including ATLAS, use a pretrained ViT-B/16~\cite{dosovitskiy2020vit}, while AVSBench, AVSBidirectional, and AVSegFormer, AVS-VCT~\cite{huang2025_visual_centric_avs} retain PVT-v2 backbones~\cite{pvtv2}. For ATLAS, we set $r=8$, $\alpha=16$, $\lambda_{cls}=0.5$, $c=0.3$. We train all methods for 30 epochs using AdamW with a learning rate $10^{-3}$ and weight decay $5\times10^{-4}$. SS-AVS follows the 11-2 split (7 tasks total), and MS-AVS uses the 31-5  split (50 tasks). Results are averaged over five runs with different seeds on four NVIDIA A40 GPUs.

\subsection{Evaluation Protocol and Metrics}
We follow the standard continual learning evaluation protocol and continual semantic segmentation metrics~\cite{gem, CL_semanticseg, lamda_pilot}. The validation set is used exclusively for model selection, while the held out test set is used to calculate the accuracy matrix $\mathbf{A}$ from which all continual learning metrics are measures. After training on task $t$, the model is evaluated on all previously seen tasks ($k \le t$) and one unseen task (when available) to measure forward transfer. Task identity is provided during evaluation in Task-Incremental Learning (TIL), while Class-Incremental, Domain-Incremental and Task-Free settings require inference without task information. 

\begin{enumerate}
\item \textbf{Segmentation Metrics:}

\begin{enumerate}
    \item   \textbf{Mean Average Precision (mAP)}
            Measures the ranking quality of confidence scores independent of a fixed threshold ($\uparrow$ better).
            \begin{equation}
            \mathrm{AP} = \sum_j \mathrm{Prec}(j),\Delta \mathrm{Rec}(j),
            \qquad
            \mathrm{mAP} = \frac{1}{|\mathcal{D}|}\sum_{i\in\mathcal{D}} \mathrm{AP}_i.
            \end{equation}
    \item \textbf{Maximum F-score (Max-F)}
            Captures the best achievable binary segmentation performance across decision thresholds ($\uparrow$ better).
            \begin{equation}
            F_1(\tau) = \frac{2P(\tau)R(\tau)}{P(\tau)+R(\tau)},
            \qquad
            \mathrm{Max\text{-}F} = \max_{\tau \in [0,1]} F_1(\tau).
            \end{equation}
\end{enumerate}
\item \textbf{Continual Learning Metrics:}

\textbf{Accuracy Matrix $\mathbf{A}$}
Let $T$ denote the total number of incremental tasks. After learning task $t$, evaluation on task $k$ yields
\begin{equation}
\mathbf{A} \in \mathbb{R}^{T \times T}, \qquad
A_{t,k} = \text{performance after task } t \text{ evaluated on task } k.
\end{equation}
This matrix summarizes performance across the task sequence.
    \begin{enumerate}
        \item \textbf{Learning Accuracy (LA)}
        Performance when each task is first learned ($\uparrow$ better).
        \begin{equation}
        \mathrm{LA} = \frac{1}{T} \sum_{t=0}^{T-1} A_{t,t}.
        \end{equation}
        \item \textbf{Average Accuracy (AA)}
        Final performance over all tasks after completing training ($\uparrow$ better).
        \begin{equation}
        \mathrm{AA} = \frac{1}{T} \sum_{k=0}^{T-1} A_{T-1,k}.
        \end{equation}
        \item \textbf{Forgetting (F)}
        Measures degradation on earlier tasks after learning new ones ($\downarrow$ better).
        \begin{equation}
        \mathrm{F} = \frac{1}{T-1} \sum_{k=0}^{T-2}
        \left( \max_{t \in {k,\dots,T-1}} A_{t,k} - A_{T-1,k} \right).
        \end{equation}
        \item \textbf{Backward Transfer (BWT)}
        Influence of learning new tasks on previously learned tasks (positive $\uparrow$ desirable).
        \begin{equation}
        \mathrm{BWT} = \frac{1}{T-1} \sum_{k=0}^{T-2}
        \left( A_{T-1,k} - A_{k,k} \right).
        \end{equation}
        \item \textbf{Forward Transfer (FWT)}
        Zero-shot generalization to future tasks before training (positive $\uparrow$ desirable).
        \begin{equation}
        \mathrm{FWT} = \frac{1}{T-1} \sum_{k=1}^{T-1} A_{k-1,k}.
        \end{equation}
    \end{enumerate}
\end{enumerate}

Additional segmentation metrics (AUPR, mIoU, and Dice) are reported in the supplementary material.

\section{Results and Discussion}
Table~\ref{tab1:main_acc} presents results across the SS-AVS benchmark (TIL, CIL, DIL; 11-2 split, 7 tasks) and MS-AVS benchmark (TF-CL; 31-5 split, 50 tasks). ATLAS achieves the highest mAP in all four settings, outperforming the runner-up by 7 to 17 points. Among exemplar-free CL methods (\cbox{cbblue}), adapted to AVS by adding segmentation decoders and a PANN~\cite{pann} audio encoder, the regularization-based EWC~\cite{EWC}, SI~\cite{synaptic_CL}, and MAS~\cite{aljundi2018memoryawaresynapseslearning} perform best on SS-AVS by penalizing changes to task-important weights via Fisher information, trajectory-based accumulation, and output sensitivity, respectively. However, their per-weight scalar importance degrades on MS-AVS with 50 tasks, where accumulated penalties conflict with large number of tasks and multiple sounding objects. L2P~\cite{L2P} steers a frozen ViT via a prompt pool, but its prompts operate only on visual tokens, leaving the audio branch with no pathway to influence the prompted representations. RanPAC~\cite{ranpac} and FeCAM~\cite{fecam} have zero forgetting by freezing features entirely, but their mAP remains near the Finetune baseline. This confirms that freezing visual features without adaptive cross-modal alignment cannot help localize sounding objects. CMR~\cite{CMR} (\cbox{Bisque2}) replays stored exemplars but achieves only 22.75 (TIL) mAP, indicating that naive replay without explicit audio-visual grounding is insufficient.

\begin{table*}[t]
\centering
\resizebox{1\columnwidth}{!}{
\begin{tabular}{p{0.5cm}|l|cc|cc|cc|cc}
\hline
               &   & \multicolumn{6}{c|}{SS-AVS (11-2 split)}                          & \multicolumn{2}{c}{MS-AVS (31-5 split)}  \\ \hline
                &  & \multicolumn{2}{c|}{TIL}                                    & \multicolumn{2}{c|}{CIL}                                    & \multicolumn{2}{c|}{DIL}               & \multicolumn{2}{c}{TF-CL}  \\ \hline
                &  & \multicolumn{1}{c}{Avg mAP} & \multicolumn{1}{c|}{Avg For} & \multicolumn{1}{c}{Avg mAP} & \multicolumn{1}{c|}{Avg For} & \multicolumn{1}{c}{Avg mAP} & \multicolumn{1}{c|}{Avg For} & \multicolumn{1}{c}{Avg mAP} & \multicolumn{1}{c}{Avg For}\\ \hline
\cellcolor{cbblue} & Finetune          & \multicolumn{1}{c}{27.91}    & \multicolumn{1}{c|}{7.85}        & \multicolumn{1}{c}{28.04}    & \multicolumn{1}{c|}{10.89}        & \multicolumn{1}{c}{25.38}        & \multicolumn{1}{c|}{9.30}         & \multicolumn{1}{c}{16.61}        & \multicolumn{1}{c}{11.58}          \\ \hline
\cellcolor{cbblue} & LwF~\cite{LWF} [TPAMI `16]& \multicolumn{1}{c}{26.70}    & \multicolumn{1}{c|}{10.34}        & \multicolumn{1}{c}{26.89}    & \multicolumn{1}{c|}{7.52}        & \multicolumn{1}{c}{24.79}        & \multicolumn{1}{c|}{12.52}         & \multicolumn{1}{c}{21.59}        & \multicolumn{1}{c}{13.43}          \\
\cellcolor{cbblue} & EWC~\cite{EWC} [PNAS `17] & \multicolumn{1}{c}{54.90}   & \multicolumn{1}{c|}{23.12}        & \multicolumn{1}{c}{55.03}    & \multicolumn{1}{c|}{23.52}        & \multicolumn{1}{c}{41.37}        & \multicolumn{1}{c|}{23.47}         & \multicolumn{1}{c}{18.89}        & \multicolumn{1}{c}{31.81}          \\
\cellcolor{cbblue} & SI~\cite{synaptic_CL} [ICML `17]  & \multicolumn{1}{c}{59.10}    & \multicolumn{1}{c|}{13.74}        & \multicolumn{1}{c}{56.52}   & \multicolumn{1}{c|}{15.57}        & \multicolumn{1}{c}{42.44}        & \multicolumn{1}{c|}{22.77}         & \multicolumn{1}{c}{19.51}        & \multicolumn{1}{c}{33.25}          \\ 
\cellcolor{cbblue} & MAS~\cite{aljundi2018memoryawaresynapseslearning} [ECCV `18]& \multicolumn{1}{c}{56.92}   & \multicolumn{1}{c|}{20.86}        & \multicolumn{1}{c}{56.39}   & \multicolumn{1}{c|}{20.23}        & \multicolumn{1}{c}{42.97}        & \multicolumn{1}{c|}{25.37}         & \multicolumn{1}{c}{26.25}        & \multicolumn{1}{c}{26.09}          \\
\cellcolor{cbblue} & L2P~\cite{L2P} [CVPR `21]& \multicolumn{1}{c}{28.80}    & \multicolumn{1}{c|}{10.59}        & \multicolumn{1}{c}{25.21}    & \multicolumn{1}{c|}{12.97}        & \multicolumn{1}{c}{23.15}        & \multicolumn{1}{c|}{9.73}         & \multicolumn{1}{c}{20.76}        & \multicolumn{1}{c}{15.87}          \\
\cellcolor{cbblue} & RanPAC~\cite{ranpac} [NIPS `23]& \multicolumn{1}{c}{29.36}    & \multicolumn{1}{c|}{0.00}        & \multicolumn{1}{c}{27.80}    & \multicolumn{1}{c|}{0.00}        & \multicolumn{1}{c}{43.99}        & \multicolumn{1}{c|}{0.00}         & \multicolumn{1}{c}{27.62}        & \multicolumn{1}{c}{0.00}          \\
\cellcolor{cbblue} & FeCAM~\cite{fecam} [NIPS `23]& \multicolumn{1}{c}{28.49}    & \multicolumn{1}{c|}{0.00}        & \multicolumn{1}{c}{28.07}    & \multicolumn{1}{c|}{0.00}        & \multicolumn{1}{c}{44.17}        & \multicolumn{1}{c|}{0.00}         & \multicolumn{1}{c}{27.33}        & \multicolumn{1}{c}{0.00}          \\
\cellcolor{cbblue} & DGR~\cite{DGR} [CVPR `24]& \multicolumn{1}{c}{31.32}    & \multicolumn{1}{c|}{40.59}        & \multicolumn{1}{c}{28.77}    & \multicolumn{1}{c|}{34.96}        & \multicolumn{1}{c}{45.92}        & \multicolumn{1}{c|}{13.09}         & \multicolumn{1}{c}{36.95}        & \multicolumn{1}{c}{19.61}          \\
\cellcolor{cbblue} & PANDA~\cite{raghavan2025pandapatchdistributionaware} [AAAI `26]& \multicolumn{1}{c}{33.13}    & \multicolumn{1}{c|}{35.34}        & \multicolumn{1}{c}{31.15}    & \multicolumn{1}{c|}{27.85}        & \multicolumn{1}{c}{46.12}        & \multicolumn{1}{c|}{12.44}         & \multicolumn{1}{c}{38.17}        & \multicolumn{1}{c}{18.03}          \\\hline
\cellcolor{cborange} & AVSBench~\cite{zhou2024avss} [ECCV `22] & \multicolumn{1}{c}{63.84}        & \multicolumn{1}{c|}{22.05}        & \multicolumn{1}{c}{62.28}        & \multicolumn{1}{c|}{23.64}        & \multicolumn{1}{c}{52.78}        & \multicolumn{1}{c|}{24.95}  & \multicolumn{1}{c}{31.15}        & \multicolumn{1}{c}{21.09}                  \\
\cellcolor{cborange} & AVS-Bidirection~\cite{AVS_birdir} [AAAI `24]    & \multicolumn{1}{c}{50.27}        & \multicolumn{1}{c|}{26.95}        & \multicolumn{1}{c}{50.49}        & \multicolumn{1}{c|}{26.19}        & \multicolumn{1}{c}{43.20}        & \multicolumn{1}{c|}{24.12}         & \multicolumn{1}{c}{31.69}        & \multicolumn{1}{c}{26.32}          \\
\cellcolor{cborange} & COMBO~\cite{combo} [CVPR `24] & \multicolumn{1}{c}{27.67}        & \multicolumn{1}{c|}{7.38}        & \multicolumn{1}{c}{27.18}        & \multicolumn{1}{c|}{8.89}        & \multicolumn{1}{c}{23.07}        & \multicolumn{1}{c|}{9.95}         & \multicolumn{1}{c}{23.76}        & \multicolumn{1}{c}{11.02}          \\ 
\cellcolor{cborange} & AVS-VCT~\cite{huang2025_visual_centric_avs} [CVPR `25] & \multicolumn{1}{c}{25.64}        & \multicolumn{1}{c|}{8.60}        & \multicolumn{1}{c}{25.54}        & \multicolumn{1}{c|}{11.18}        & \multicolumn{1}{c}{22.19}        & \multicolumn{1}{c|}{13.49}         & \multicolumn{1}{c}{21.98}        & \multicolumn{1}{c}{13.87}          \\ \hline
\cellcolor{cbpurple} & BalConpas~\cite{chen2024strike} [ECCV `24] & \multicolumn{1}{c}{23.71}        & \multicolumn{1}{c|}{10.84}        & \multicolumn{1}{c}{22.90}        & \multicolumn{1}{c|}{10.07}        & \multicolumn{1}{c}{21.68}        & \multicolumn{1}{c|}{11.44}         & \multicolumn{1}{c}{20.16}        & \multicolumn{1}{c}{16.92}          \\ 
\cellcolor{cbpurple} & EIR~\cite{EIR_2025} [CVPR `25] & \multicolumn{1}{c}{29.54}        & \multicolumn{1}{c|}{9.12 }        & \multicolumn{1}{c}{28.91}        & \multicolumn{1}{c|}{10.44}        & \multicolumn{1}{c}{26.70}        & \multicolumn{1}{c|}{9.73}         & \multicolumn{1}{c}{22.77}        & \multicolumn{1}{c}{10.71}          \\ 
\cellcolor{cbpurple} & CSTA$^{*}$~\cite{chen2025csta} [TCSV `25]  & \multicolumn{1}{c}{20.68}        & \multicolumn{1}{c|}{1.31}        & \multicolumn{1}{c}{20.41}        & \multicolumn{1}{c|}{0.24}        & \multicolumn{1}{c}{19.32}        & \multicolumn{1}{c|}{0.16}         & \multicolumn{1}{c}{14.50}        & \multicolumn{1}{c}{0.60}          \\
\cellcolor{cbpurple} & HVPL$^{*}$~\cite{Dong2025HVPL} [ICCV `25] & \multicolumn{1}{c}{28.31}        & \multicolumn{1}{c|}{0.98}        & \multicolumn{1}{c}{32.68}        & \multicolumn{1}{c|}{19.04}        & \multicolumn{1}{c}{35.97}        & \multicolumn{1}{c|}{1.56}         & \multicolumn{1}{c}{26.50}        & \multicolumn{1}{c}{0.26}          \\ \hline
\cellcolor{Bisque2} & CMR$^{*}$~\cite{CMR} [arXiV `25] & \multicolumn{1}{c}{22.75}        & \multicolumn{1}{c|}{0.32}        & \multicolumn{1}{c}{19.16}        & \multicolumn{1}{c|}{2.77}        & \multicolumn{1}{c}{19.59}        & \multicolumn{1}{c|}{1.30}         & \multicolumn{1}{c}{25.92}        & \multicolumn{1}{c}{1.35}          \\ \hline \hline
\cellcolor{cbgreen} & ATLAS\textit{(Ours)} (\textit{Ours})              & \multicolumn{1}{c}{\textbf{74.67}}        & \multicolumn{1}{c|}{9.42}        & \multicolumn{1}{c}{\textbf{71.39}}        & \multicolumn{1}{c|}{10.14}        & \multicolumn{1}{c}{\textbf{63.84}}        & \multicolumn{1}{c|}{13.96}         & \multicolumn{1}{c}{\textbf{45.27}}        & \multicolumn{1}{c}{23.71}          \\ \hline
\end{tabular}
}
\caption{\textbf{Performance on the Continual AVS Benchmark} \cbox{cbblue} exemplar-free CL methods adapted to AVS; \cbox{cborange} static AVS models extended to continual learning; \cbox{cbpurple} continual segmentation adapted to AVS; \cbox{Bisque2} the CL-AVS baseline with replay; and \cbox{cbgreen} our method. Performance is measured by mAP across tasks and average backward forgetting.}
\label{tab1:main_acc}
\end{table*}

\begin{figure}[h!]
    \centering
    \includegraphics[width=0.85\linewidth]{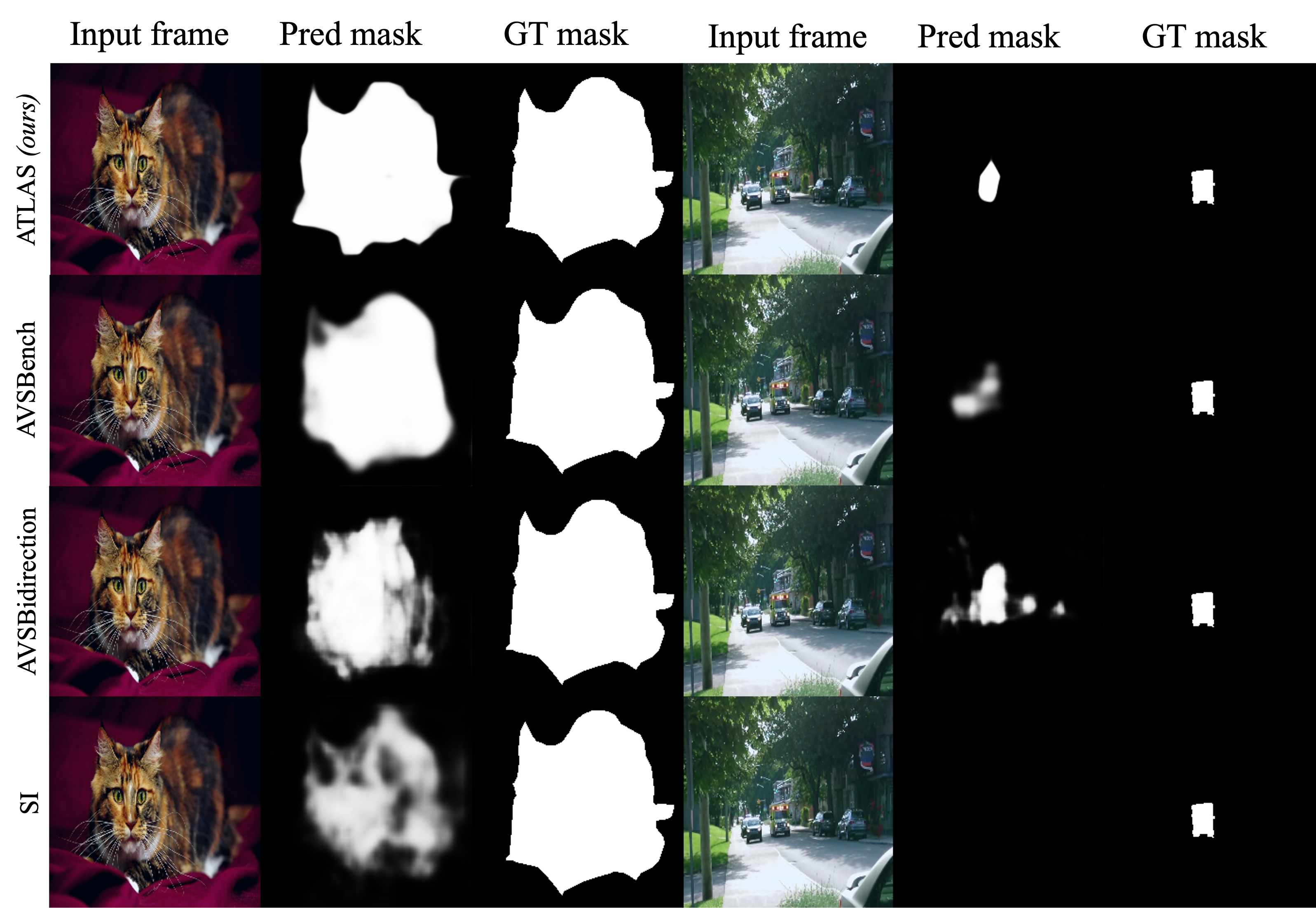}
    \caption{\textbf{Qualitative AVS results from the top four methods in our CL-AVS setting}: Input frame, predicted binary mask for the middle frame is shown along with the ground truth mask. \textbf{(Left)} SS-AVS CIL and \textbf{(Right)} MS-AVS TF-CL}
    \label{fig:segment_visuals}
\end{figure}

Among static AVS models (\cbox{cborange}) trained continually, AVSBench~\cite{zhou2024avss} performs best on PVT-v2 due to its dedicated audio-visual interaction module, but lacks a CL mechanism and suffers from high forgetting as continuous fine-tuning overwrites learned fusion weights. AVS-Bidirection~\cite{AVS_birdir} performs worse due to bidirectional attention, as parameter updates go through both visual and audio directions and increase forgetting. COMBO~\cite{combo} and AVS-VCT~\cite{huang2025_visual_centric_avs} hover near the Finetune baseline, suggesting their single-task design improvements lead to no continual learning improvements. Continual segmentation methods (\cbox{cbpurple}) adapted with audio encoders and MSE for alignment exhibit a pattern. HVPL~\cite{Dong2025HVPL} and CSTA~\cite{chen2025csta} achieve very low forgetting but fair mAP, as their single modality design updates too few parameters to learn the cross-modal correspondences AVS needs. DGR~\cite{DGR} and PANDA~\cite{raghavan2025pandapatchdistributionaware} perform well through gradient reweighting and patch-distribution mechanisms. Their high forgetting rates reveal their shortcomings in adapting to two modalities. We attribute the gains of our proposed model, ATLAS, to our major components. Firstly, LoRA adapters keep updates within the low-rank subspaces of the visual encoder and decoder. Followed by audio-guided pre-fusion conditioning that leads visual features toward sound-relevant regions before performing cross-attention. Finally, our Low-Rank Anchoring dynamically penalizes adapter and decoder weight drift from previous-task anchors. Additionally, Figure~\ref{fig:segment_visuals} shows qualitative segmentation results for the four best-performing methods, highlighting the strength of our approach. 

\begin{figure}[h!]
    \centering
    \includegraphics[width=1\linewidth]{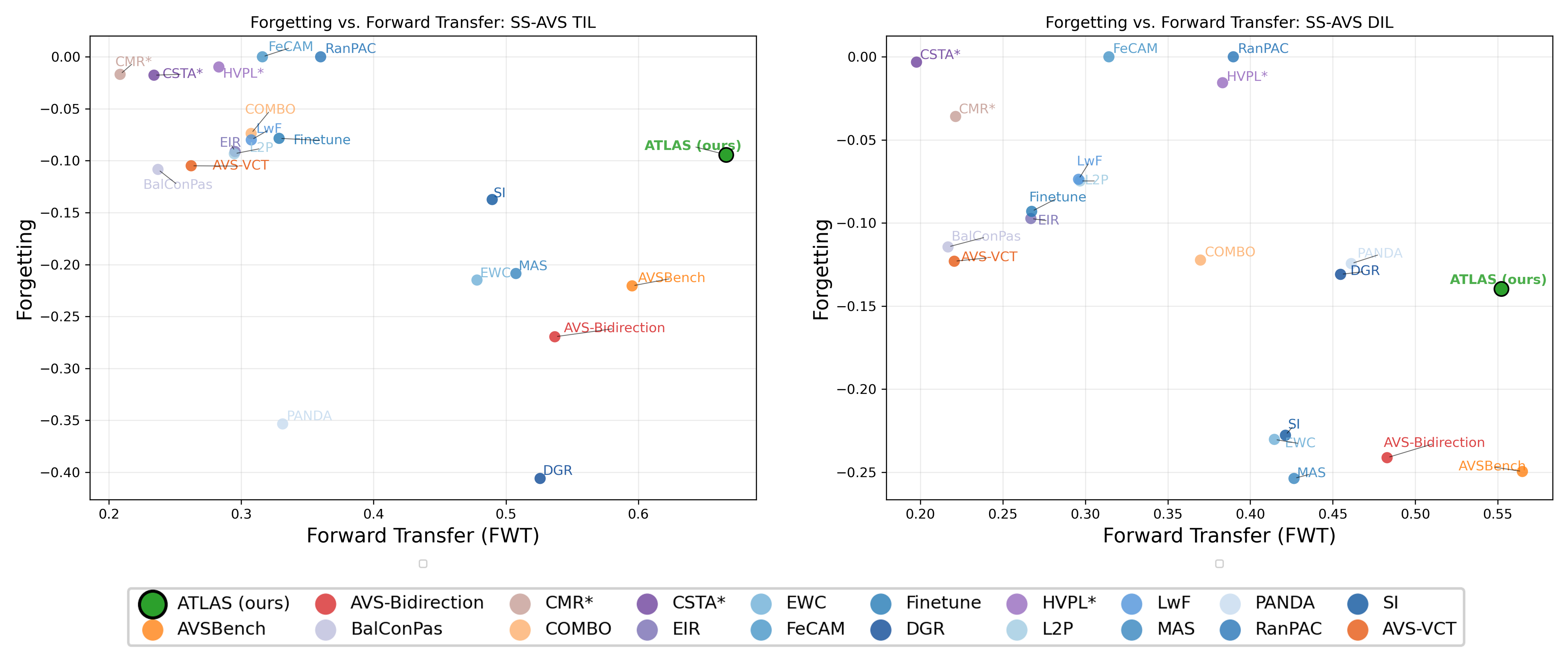}
    \caption{\textbf{Forward Transfer vs. Forgetting:} Trade-off between forward transfer and forgetting across models in the CL-AVS benchmark on the SS-AVS dataset under the TIL and DIL protocols.}
    \label{fig:fwt-for}
\end{figure}

Figure~\ref{fig:fwt-for} plots Forward Transfer vs.\ Forgetting for TIL and DIL in the SS-AVS dataset with 11-2 split. ATLAS achieves the highest forward transfer with moderate forgetting. Most methods cluster at low FWT ($<0.45$), either forgetting heavily (DGR, PANDA, MAS) or preserving knowledge at the cost of plasticity (CSTA, FeCAM, RanPAC). Under DIL, the gap is significant. ATLAS exceeds 0.55 FWT while maintaining comparable forgetting rates to those of lower-performing models. 

\begin{figure}[h!]
    \centering
    \includegraphics[width=1\linewidth]{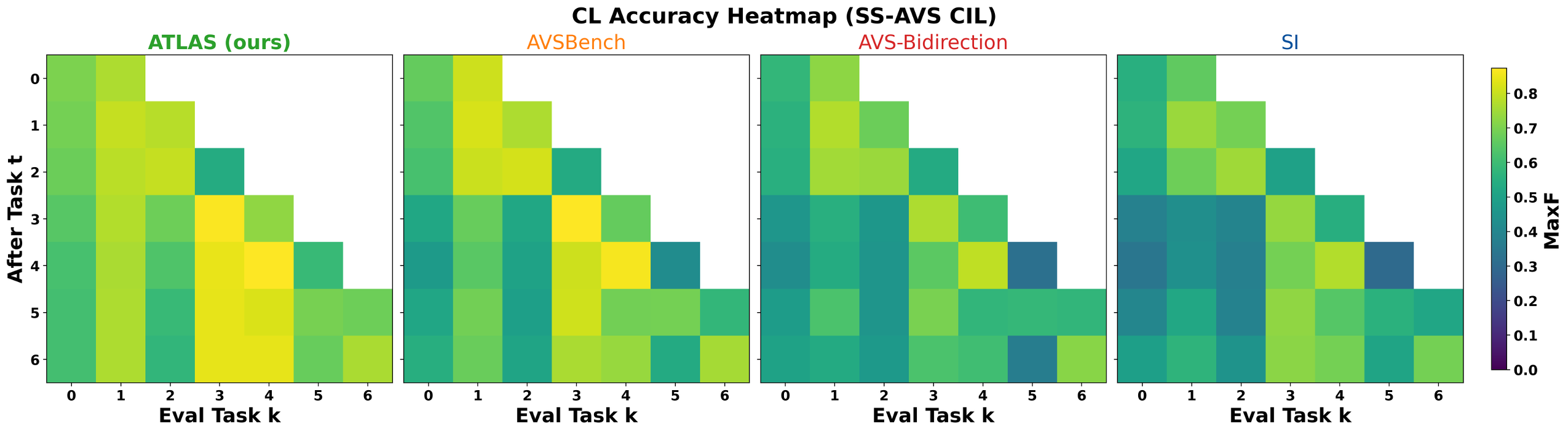}
    \caption{\textbf{Heatmap of MaxF scores} on the SS-AVS test set under the CIL protocol with an 11–2 split, evaluated after each incremental task. Four best-performing methods are included for clarity.}
    \label{fig:maxF_heatmap}
\end{figure}

Figure~\ref{fig:maxF_heatmap} shows MaxF heatmaps for the CIL setting in SS-AVS. ATLAS maintains bright cells across tasks, with early tasks remaining strong even by row 6, suggesting that our Low-Rank Anchoring mechanism effectively limits drift from previously learned representations. AVSBench performs well on early tasks but gradually reduces as new tasks are introduced since it is not designed for continual learning. AVS-Bidirectional shows a sharp drop after task 3, likely due to error propagating through its bidirectional interactions. In contrast, SI preserves individual parameter importance but struggles to maintain the distributed cross-modal representations required for accurate segmentation. Further results and visualizations are provided in the supplementary material, including hyperparameter analyzes, additional metrics, and experiments with alternative backbone architectures. We note that the performance of the continual learner depends strongly on the quality of the pretrained models.

\section{Ablation Studies}

\begin{table}[h!]
\centering
\resizebox{1\columnwidth}{!}{
\begin{tabular}{lcc|cccc|cccc}
\hline
 &  &  & \multicolumn{4}{c|}{SS-AVS (11-2 split)} & \multicolumn{4}{c}{MS-AVS (31-5 split)} \\
\cline{4-11}
Method & Pre-cond. & LRA 
& Avg mAP & Avg For & Avg F1 & Avg mIoU
& Avg mAP & Avg For & Avg F1 & Avg mIoU \\
\hline

ATLAS & \cmark & \cmark & \textbf{74.67} & \textbf{9.42}  &\textbf{62.76}  &\textbf{50.85}  &\textbf{47.53}  & 23.71 &\textbf{42.27}  &\textbf{39.44}  \\
ATLAS & \xmark & \cmark & 71.17  & 12.11  & 59.97  & 47.78  & 43.12  & 25.07  &39.38  &38.2 7  \\
ATLAS & \cmark & \xmark & 67.18  & 18.03  & 50.38  & 39.16  & 34.41  &26.61   &36.72  &24.12  \\
ATLAS & \xmark & \xmark & 65.33  & 21.00  & 47.31  &33.58  &29.19  &27.01  &31.16  &20.12  \\ \hline
AVSBench & - & - &63.84 &22.05 &50.63 &37.28 &31.15 &\textbf{21.09} &34.39 &20.15 \\ \bottomrule
\end{tabular}
}
\caption{\textbf{Component ablation of ATLAS.} 
We analyze the contribution of audio-guided pre-conditioning and Low-Rank Anchoring (LRA) on the CL-AVS benchmark}
\label{tab:abl}
\end{table}

\begin{figure}[h!]
    \centering
    \includegraphics[width=1\linewidth]{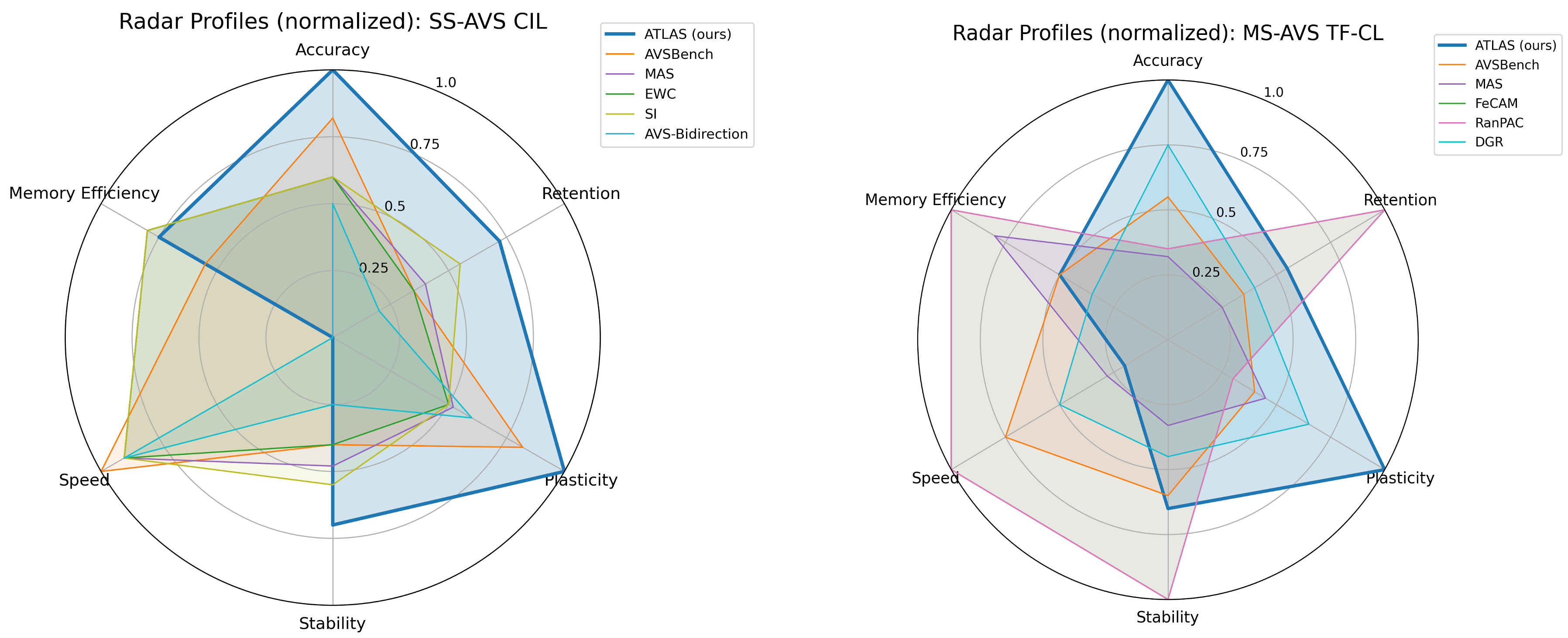}
    \caption{\textbf{Radar plot} summarizing normalized performance across six dimensions: Accuracy, Retention, Plasticity, Stability, Speed, and Memory Efficiency for the top six performing methods on the SS-AVS dataset under the CIL protocol and the MS-AVS dataset under the TF-CL protocol.}
    \label{fig:radarplot}
\end{figure}
In this section, we conduct ablation studies analyzing efficiency–performance trade-offs and the contributions of individual components of our method. Figure~\ref{fig:radarplot} shows normalized radar profiles across six metrics: Accuracy, Retention, Plasticity, Stability, Speed, and Memory Efficiency. ATLAS covers the largest area for both SS-AVS (CIL) and MS-AVS (TF-CL), showing strong performance particularly in Accuracy and Plasticity. Regularization methods exhibit smaller overall coverage, while prototype based approaches such as RanPAC and FeCAM are memory efficient but tend to underfit. 
To understand the contribution of each component of ATLAS, we perform ablations summarized in Table~\ref{tab:abl}, using AVSBench as a baseline for comparison. Results show that Low-Rank Anchoring (LRA) is the most critical component, as it stabilizes LoRA-adapted weights and reduces parameter drift during continual learning. Audio-guided pre-fusion conditioning provides additional gains but contributes less than LRA. Even without these modules, the LoRA-based model suffers from parameter drift across tasks, resulting in performance below AVSBench primarily on MS-AVS larger number of tasks.

\section{Conclusion}
This paper introduces an exemplar-free continual learning benchmark for audio-visual segmentation, aimed at producing pixel-level masks of sounding objects in videos. We evaluate a diverse set of methods across four continual learning protocols on both Single-Source and Multi-Source AVS datasets. We further propose ATLAS, an end-to-end exemplar-free baseline that improves learning while reducing representation drift. Our experiments demonstrate its effectiveness and establish a foundation for future research in continual audio-visual learning.

%


\clearpage
%
%
\bibliographystyle{splncs04}
\bibliography{main}
\end{document}